\documentclass[10.5pt]{article}
\PassOptionsToPackage{numbers, compress}{natbib}
\usepackage[utf8]{inputenc}
\usepackage{amsmath}
\usepackage{amsfonts}
\usepackage{mathrsfs}
\usepackage{amssymb}
\usepackage{amsthm}

\usepackage{libertine}
\usepackage{wrapfig}
\usepackage{graphicx}
\usepackage{multirow}
\usepackage{epstopdf}
\usepackage{multicol}
\usepackage{booktabs}

\usepackage{subfig}

\usepackage{algorithmic}
\usepackage{algorithm}
\usepackage[algo2e,linesnumbered]{algorithm2e}
\usepackage{stackengine}
\usepackage{graphicx}
\usepackage{multirow}
\usepackage{epstopdf}
\usepackage{multicol}
\usepackage{caption}
\usepackage{enumitem}
\usepackage{amsmath}
\usepackage{pifont}
\usepackage{natbib}
\usepackage{bm}
\usepackage{bbm}
\usepackage{makecell}
\usepackage{mathtools}
\usepackage[table]{xcolor}
\definecolor{mydarkred}{rgb}{0.6,0,0}
\definecolor{mydarkgreen}{rgb}{0,0.6,0}
\usepackage[colorlinks,
linkcolor=mydarkred,
citecolor=mydarkgreen]{hyperref}
\usepackage{url}
\usepackage{bm}
\usepackage{pifont}
\usepackage{stfloats}
\usepackage[T1]{fontenc}
\usepackage{arydshln}
\usepackage{colortbl}
\usepackage{balance}

\newcolumntype{L}[1]{>{\raggedright\let\newline\\\arraybackslash\hspace{0pt}}m{#1}}
\newcolumntype{Y}{>{\centering\arraybackslash}X}
\newcolumntype{s}{>{\hsize=.3\hsize}Y}
\newcolumntype{t}{>{\hsize=1.5\hsize}X}
\newcolumntype{u}{>{\hsize=0.8\hsize}Y}
\usepackage{makecell}
\usepackage{mathtools}
\usepackage[utf8]{inputenc}
\usepackage{stfloats}

\title{MSR: Making Self-supervised learning Robust to \\
Aggressive Augmentations}

\author{
  Yingbin Bai$^{1}$,  
  Erkun Yang$^{2}$,   
  Zhaoqing Wang$^{1}$,
  Yuxuan Du$^3$, \\
  Bo Han$^4$,
  Cheng Deng$^{2}$,
  Dadong Wang$^{5}$,
  Tongliang Liu$^1$\thanks{Correspondence to Tongliang Liu (tongliang.liu@sydney.edu.au)}\\[1ex]
  \small{$^1$TML Lab, The University of Sydney;}
  \small{$^2$Xidian University;}\\
  \small{$^3$JD Explore Academy;} 
  \small{$^4$Hong Kong Baptist University}
  \small{$^5$CSIRO}
}
\date{}

\begin{document}
\maketitle

\begin{abstract}
Most recent self-supervised learning methods learn visual representation by contrasting different augmented views of images. Compared with supervised learning, more aggressive augmentations have been introduced to further improve the diversity of training pairs. However, aggressive augmentations may distort images’ structures leading to a severe semantic shift problem that augmented views of the same image may not share the same semantics, thus degrading the transfer performance. To address this problem, we propose a new SSL paradigm, which counteracts the impact of semantic shift by balancing the role of weak and aggressively augmented pairs. Specifically, semantically inconsistent pairs are of minority and we treat them as noisy pairs. Note that deep neural networks (DNNs) have a crucial memorization effect that DNNs tend to first memorize clean (majority) examples before overfitting to noisy (minority) examples. Therefore, we set a relatively large weight for aggressively augmented data pairs at the early learning stage. With the training going on, the model begins to overfit noisy pairs. Accordingly, we gradually reduce the weights of aggressively augmented pairs. In doing so, our method can better embrace the aggressive augmentations and neutralize the semantic shift problem. Experiments show that our model achieves 73.1\% top-1 accuracy on ImageNet-1K with ResNet-50 for 200 epochs, which is a 2.5\% improvement over BYOL. Moreover, experiments also demonstrate that the learned representations can transfer well for various downstream tasks. 
\end{abstract}

\section{Introduction}
A golden law in the context of computer vision is utilizing tremendous annotated data to learn good visual representations \cite{Daokun2020RepreSurvey, Phuc2020ContrastSurvey}. Unfortunately, collecting annotated data with accurate labels is generally laborious, expensive \cite{Yan2014annotators, Xi2018Complementary}, and even infeasible \cite{Davood2020medical}. To this end, various approaches have been proposed to learn such representations from unlabeled visual data, usually by performing visual pretext tasks. Among them, self-supervised learning methods \cite{Chen2020simclr, he2020mocov2, Wang2020uniformity, chuang2020debiased} based on contrastive loss have recently shown great promise, achieving state-of-the-art performance.

\begin{figure}[!t]
\centering
\hspace*{\fill}%
\subfloat[Noisy samples from ImageNet-1K \cite{krizhevsky2012imagenet}]{\includegraphics[width=0.49\textwidth]{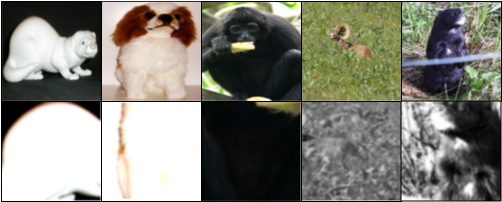}}
\hfill
\subfloat[Noisy samples from CIFAR-100 \cite{krizhevsky2009CIFAR}]{\includegraphics[width=0.49\textwidth]{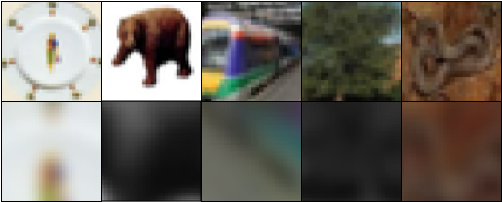}} 
\hspace*{\fill}
\caption{The first row of Figure 1 are the original images (ImageNet-1K images are resized to squares) and noisy samples from aggressive augmentation are in the second row. From the first three images of (a), we can observe that Color jitter operation makes the image too bright or too dark that covers the details of images; and in the fourth and fifth column of (a), Gray and Gaussian blur operation leads images to be hardly distinguished from background; and the same augmentation strategy from ImageNet-1K leads to more vague images for CIFAR-100 shown in (b).}
\label{fig:noisy}
\vspace{-5px}
\end{figure}

Representative contrastive methods are generally trained by maximizing agreement between differently augmented views of the same image (positive pairs), and increasing the distance between augmented views from different images (negative pairs) \cite{Wu2018instance, Chen2020simclr, he2020moco}. Compared with supervised learning, these works highlight the role of data augmentation for SSL and design more aggressive augmentation operations, such as grayscale, color jitter, and Gaussian blur. Although these aggressive augmentations can help to further improve the model performance, they also bring 
a severe semantic shift problem for training images. As illustrated in Figure \ref{fig:noisy}, the first row shows original images from ImageNet \cite{krizhevsky2012imagenet} and CIFAR-100 \cite{krizhevsky2009CIFAR} datasets. And the second row presents the corresponding augmented views with the widely used composition of augmentations \cite{he2020mocov2, Chen2020simclr}. We can see that the augmented views can be hardly recognized as semantically consistent with their original versions.  Pushing these images to have similar representations can adversely affect the model training.  Some recent works \cite{Mingkai2021ressl, Soroush20P21meanshift} have also recognized this problem and resort to using weak augmentation to avoid it. However, directly discarding aggressive augmentations may reduce the diversities of training examples, resulting in limited representation ability. Therefore, in this paper, we retain the aggressive augmentations and try to counteract the subsequent semantic shift problem.  

Specifically, we consider the semantically inconsistent pairs as noisy positive pairs. Since they are mixed with other semantic consistent pairs, it is hard to directly filter them out during training. Fortunately, recent works \cite{Arpit17CloserLook, han2018co, Zhang2017rethinking} show that deep neural networks (DNNs) have  a crucial memorization effect that DNNs tend to first memorize clean (majority/semantically consistent) examples before overfitting noisy (minority/semantically inconsistent) examples. 
Motivated by this, we propose to set a relatively large weight for the aggressively augmented data pairs at the beginning of training to fully exploit all the training examples. And as the training goes on, the model begins to overfit semantically inconsistent data. Therefore we gradually reduce the weight of aggressive augmented pairs to neutralize their impact. Compared with ReSSL \cite{Mingkai2021ressl}, our method can embrace the diverse examples from aggressive augmentations. And, compared with MoCo \cite{he2020moco}, SimCLR \cite{Chen2020simclr}, BYOL \cite{Grill2020byol}, and few others, our method can significantly reduce the semantic shift problem.

Experiment results on multiple benchmark datasets  show that our method can outperform state-of-the-art methods in various settings with a large margin. For instance, with 200 epochs of pre-training, our method achieves 73.1\% Top-1 accuracy on ImagetNet-1K \cite{krizhevsky2012imagenet} linear evaluation protocol, which is 2.5\% higher than BYOL \cite{Grill2020byol}. Experiments on MS COCO \cite{Lin2014COCO} also show that our pre-trained model can continually improve the performance for multiple downstream tasks.

\section{Related works}

\textbf{Self-supervised learning (SSL)} has attracted great attention to capture universal representations \cite{hwang2019segsort, zhang2020hierarchical, o2020dense, wang2022exploring, Caron2020SwAV, Ermolov2021whiten, Mingkai2021ressl}. The core of SSL is designing agent tasks, which allow us to learn representations from large-scale unlabeled data via pseudo labels instead of using any human annotations. To this end, many proposals devise different solutions in constructing pseudo labels, including predicting the rotation of images \cite{Spyros2018Rotation}, putting pieces of images together \cite{Noroozi2016Jigsaw}, or recovering color from grayscale images \cite{Richard2016Colorization}. Particularly, Wu et al. \cite{Wu2018instance} propose an instance-level classification, which regards images augmented from the same image as a positive pair and others as negative examples. SimCLR \cite{Chen2020simclr} improves performance by inserting the projection network and introducing aggressive augmentation. He et al \cite{he2020moco, he2020mocov2} store negative representations in a queue to reduce the memory requirement. BYOL \cite{Grill2020byol} enhances the power of SSL by removing the dependence of negative examples, which also addresses the problem of false negative examples \cite{chuang2020debiased}. Despite these methods have proved their effectiveness based on the aggressive augmentation, they ignore the semantic shift problem from it.

\textbf{Noisy samples in aggressive augmentation}. Recent studies \cite{Wang2021uota, Selvaraju2021casting, Sangwoo2021ObjectAware, peng2022crafting} have discovered that aggressive augmentation may generate noisy samples in the positive pairs. To alleviate this issue, ContraCAM \cite{Sangwoo2021ObjectAware} proposes a two-step approach to reduce the issue of random cropping, which seeks objects first and then crops images based on their locations. In addition, Gansbeke et al. \cite{Gansbeke2021revisiting} conduct experiments on scene-centric datasets (e.g., COCO) containing multiple objects in images and argue that SSL can overcome the issue of random cropping. Unlike prior works focusing on random cropping, our work allows more general types of augmentation and can be viewed as a complement of previous studies.

\textbf{Learning with noisy labels.} Memorization effect of deep neural networks has been widely studied in the field of learning with noisy labels \cite{Arpit17CloserLook, Zhang2017rethinking}. In this regard, many state-of-the-art methods select confident examples by using first fit examples in the early learning phase \cite{Liu2020ELR, Northcutt2017confident, Jiang18MentorNet, Nguyen2020SELF, Li2020DivideMix}. Specifically, Co-teaching \cite{han2018co} uses the small-loss strategy to choose confident examples and employs two networks to reduce the bias from examples with noisy labels. JointOptim \cite{tanaka2018joint} selects confident examples based on the early stopping and replaces the labels of low confident examples to predictions. DivideMix \cite{Li2020DivideMix} further improves Co-teaching \cite{han2018co} by regarding low confident examples as unlabeled data and training with semi-supervised learning techniques \cite{Berthelot2019MixMatch}. Although our method also leverages the memorization effect, existing works in learning with noisy labels focus on addressing large noise rate issues, which means that noise can be separated from training data by using early stopping trick. However, the noise rate in aggressive augmentation is very low, so we propose an indirect approach that does not extract noise to counter the noise impacts.

\section{Methodology}
Our approach aims to minimize the negative impacts of noisy positive pairs from aggressive augmentations while taking advantage of aggressive augmentations. As such, we first revisit preliminaries on self-supervised learning. Then, we elaborate on the proposed learning algorithm that counterbalances the noise impacts by utilizes memorization effects of DNNs.

\subsection{Preliminaries on self-supervised learning}

Self-supervised learning methods based on contrastive learning generally requires learning an embedding space that can easily separate different examples. Assuming $D$ be a distribution of an image set. An image $x$ is uniformly drawn from $D$. Denote $t$ and $t'$ as two different instances from the same distribution of image augmentation $\mathrm{T}$. $v$ and $v'$ are two augmented views of the image $x$ with $v = t(x)$  and $v' = t'(x) $, which are regarded as a pair of positive examples. Then, $v$ and $v'$ will separately feed through a encoder $f_\theta$ and a projector $g_\theta$ to embed $z_1$ and $z_2$, which are required to be close to each other via a contrastive loss function, e.g., InfoNCE \cite{van2018cpc} can be expressed as,

\begin{equation}
\label{eq:nce_loss}
\begin{aligned}
\mathcal{L}_{NCE} = - \log \frac{\exp(z_1 \cdot z_2/\gamma)}{\exp (z_1 \cdot z_2/\gamma) + \sum_{n \in N} \exp(z_1 \cdot n / \gamma)},
\end{aligned}
\end{equation}

where $\gamma$ is a temperature parameter, and N is a set of negative example vectors. The embeddings of positive and negative examples are $l_2$-normalized. To stabilize the training process, some state-of-the-art methods \cite{he2020moco, Mingkai2021ressl, Soroush20P21meanshift} employ an asymmetric framework, including an online and a target networks. For the online and target networks, they have the same network structure but different weights of encoder $f_\xi$, and projector $g_\xi$, whose parameters $\xi$ are updated by the online parameters $\theta$ with the exponential moving average method. 

Recently, BYOL \cite{Grill2020byol} finds negative examples are not necessary and adds a predictor $q_\theta$ in the online network to avoid collapsed solutions, e.g., all images have the same vector. And, the loss function can be simplified to,

\begin{equation}
\label{eq:mse_loss}
\begin{aligned}
\mathcal{L}_{mse} &= 2 - 2 * \frac{\langle z_{1}, z_{2} \rangle}{\lVert z_{1} \rVert_2 * \lVert z_{2} \rVert_2}.
\end{aligned}
\end{equation}

\subsection{Description of MSR}

\begin{figure}[!t]
\centering
\includegraphics[width=1\textwidth]{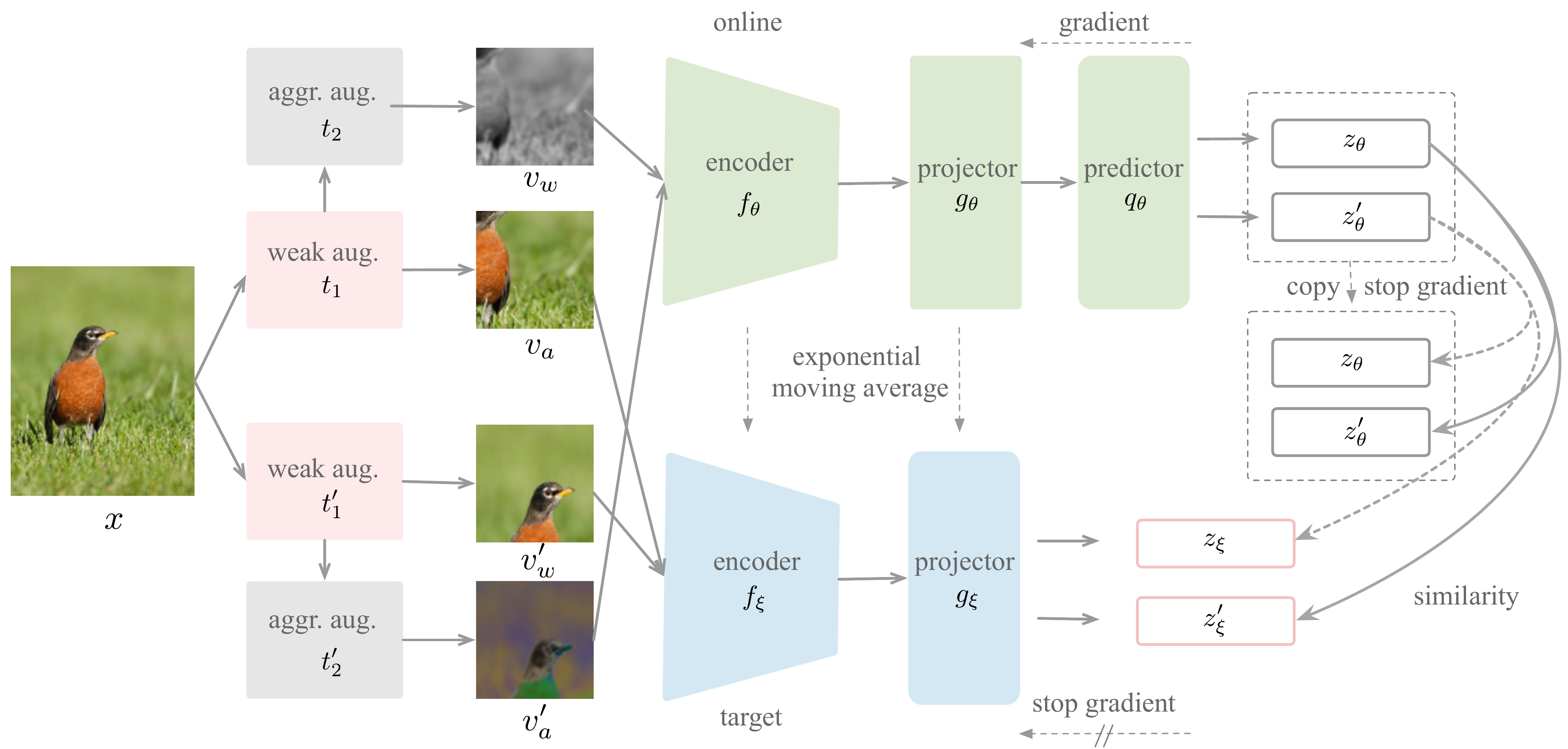}
\caption{The illustration of our proposed method (MSR). We utilize an asymmetric-style framework, including an online and a target networks. The online network is optimized by gradients, and the target network is updated with the exponential moving average strategy. We first adopt the weak augmentation to generate two views ($v_w$, $v_w'$), then adopt the aggressive augmentations to further generate another two views ($v_s$, $v_s'$). Subsequently, we make aggressive-augmented views to keep consistent with their corresponding weak- and aggressive-augmented views in the embedding space. (Best viewed in color)} 
\label{fig:arch}
\vspace{-5pt}
\end{figure}

As discussed in Introduction, noisy positive pairs from aggressive augmentations will lead to severe semantic shift problem, which damages the generalization on downstream tasks. However, most of sample pairs generated from aggressive augmentations are beneficial for the model performance and what precise degree of distortion would cause bias to remain unknown, so it is hard to distinguish from clean pairs and noisy pairs in SSL. 

To this end, we propose an efficient method making self-supervised learning robust to the noise from aggressive augmentations, dubbed MSR. Envisioned by the memorization effects, DNNs will first fit the majority of training data in the early learning phase, and then overfit to the minority. Noisy pairs account for the minority, so we assume that the noise impacts will be small and then increases as the training process. We give different weights to clean and noisy pairs, and gradually reduce weights of noisy pairs to against raised noise impacts. Instead of separating noise from training data, we employ weak augmentation to generate sample pairs as relative clean pairs.

However, simply introducing clean pairs will significantly increase computation, and SSL has known that more computation will increases performance, which makes it difficult to fairly compare with other baselines. To limit computation, we meet two problems. First, generating more augmented instances, especially double instances, commonly means expensive computation, largely burdening CPU and hard disk, which leads to slow data processing down and low efficiency of GPU \cite{Chen2020simclr}. Second, more instances require more back-propagation times, which results in more processing time for GPU.

To address the first problem, we propose a novel data augmentation pipeline, called multi-stage augmentation, which can generate double instances with nearly the same resources. Specifically, data augmentation generates image variants by mixing different types of image transformation together and arranging into a queue. For an augmentation, it receives the image from the result of the previous augmentation, and transforms the image based on pre-defined probability, and then pass it to the next transformation, $t_{aug}(x) = t_2 (t_1(x))$, where $t_1$ and $t_2$ are subset of $t_{aug}$. Based on this process, we can easily separate it into two child processes. Namely, $t_1$ includes weak augmentation operations while $t_2$ conducts aggressive augmentation operations based on the result of $t_1$. This process can be described as,
\begin{equation}
\label{eq:4aug2}
\begin{aligned}
v_{w}&=t_{1}(x_i)   & v_{w}' &= t_{1}' (x_i), \\ 
v_{a}&=t_{2}(v_{w}) & v_{a}' &= t_{2}' (v_{w}').
\end{aligned}
\end{equation}

The noisy samples will have different impacts when occurring on different networks because of the asymmetric-style framework. When noisy samples occur on the online network and the target network receives clean samples, the target representation will help correct the online representation. If noisy samples occur on the target network, the online network will receive the wrong direction signal whether it receives clean samples or not. Therefore, we send aggressive augmented views into the online network and send weak augmented views into the target network. 

\begin{equation}
\label{eq:4aug_feature}
\begin{aligned}
z_{\theta}  &= q_{\theta}(g_{\theta}(f_{\theta} (v_{a})))& z_{\xi}' &= g_{\xi}(f_{\xi} (v_{w}')), \\
z_{\theta}' &= q_{\theta}(g_{\theta}(f_{\theta} (v_{a}'))) & z_{\xi}  &= g_{\xi}(f_{\xi} (v_{w})).
\end{aligned}
\end{equation}

After obtaining four representations, we group them into four pairs, shown in Figure \ref{fig:arch} and make each aggressive-augmented view to keep consistent with their corresponding weak- and aggressive-augmented views in the embedding space. For the second problem, instead of encoding aggressive augmented views on the target network, we use representations from the predictor due to the similarity of both. Accordingly, MSR only requires back-propagation twice, which is equal to many state-of-the-art SSL methods with a symmetrized loss \cite{Chen2020simclr,he2021simsiam, Grill2020byol, Caron2020SwAV}. The total loss is summarized as,

\begin{equation}
\label{eq:loss}
\begin{aligned}
\mathcal{L}_{total} =  (1 - \beta ) \mathcal{L}_{mse}(z_{\theta}, z'_{\xi})
+ \beta  \mathcal{L}_{mse}(z_{\theta}, z'_{\theta})
+  (1 - \beta )  \mathcal{L}_{mse}(z'_{\theta}, z_{\xi})
+ \beta  \mathcal{L}_{mse}(z'_{\theta}, z_{\theta}),
\end{aligned}
\end{equation}

where $\beta$ is a calculated parameter to re-weight the losses from weak and aggressive augmented samples. This loss function forces networks to learn representation that achieve a balance between weak- and aggressive-augmented views, which reduces overfitting to noisy pairs while avoiding a simple solution. To further against the raised noise impacts and increase the contribution of weak augmented samples, we decrease $\beta$ with a cosine decay equation, 

\begin{equation}
\label{eq:beta}
\begin{aligned}
\beta = \beta_{base} \times \frac{1}{2} (cos(\pi \frac{k}{K})+1),
\end{aligned}
\end{equation}

where $\beta_{base}$ is a given number at the training beginning, and k and K are the current training steps and the total training steps.

\begin{algorithm}[!tp]
{\bfseries Input}:  Neural network $f_{\theta}$ and $f_{\xi}$. Projector $g_{\theta}$ and $g_{\xi}$. Predictor $q_{\theta}$. A batch of samples $x$. An image set $D$. Total number of training steps $K$. Weak augmentation function $t_1$. Aggressive augmentation function $t_2$. Hyper-parameter $\beta_{base}$ in \ref{eq:loss}. 

\For{$k = 1, \dots, K$}{
     $x$ is drawn from $D$ \\
       $v_{w}=t_{1}(x)  \ \ \ v_{w}' = t_{1}' (x)$ \\ 
       $v_{a}=t_{2}(v_{w}) \ \ v_{a}' = t_{2}' (v_{w}')$ \\
     \texttt{\\}
     $z_{\theta} = q_{\theta}(g_{\theta}(f_{\theta} (v_{a}))) \ \ z_{\xi}'=g_{\xi}(f_{\xi} (v_{w}'))$\\ 
     $z_{\theta}' = q_{\theta}(g_{\theta}(f_{\theta} (v_{a}'))) \ \ z_{\xi}=g_{\xi}(f_{\xi} (v_{w}))$ \\
     \texttt{\\}
     Update $f_{\theta}$, $g_{\theta}$ and $q_{\theta}$ with loss function \ref{eq:mse_loss} and \ref{eq:loss}. \\
     Update $f_{\xi}$ and $g_{\xi}$ by slowly momentum update with the parameters of  $f_{\theta}$ and $g_{\theta}$ \\
     Update $\beta$ with equation \ref{eq:beta}
}
{\bfseries Output}: The trained network $f_{\theta}$ \\
\caption{MSR: Making Self-supervised learning Robust}
\label{algorithm1}
\end{algorithm}

\section{Experiments}

\subsection{Datasets and Implementation details}
\label{sec:impl}
\textbf{Datasets:} We evaluate our method on six image datasets, from small to large. We choose CIFAR-10/100 \cite{krizhevsky2009CIFAR} for small datasets, and STL-10 \cite{Adam2011stl10} and Tiny ImageNet \cite{Zoheb2019tinyimagenet} for medium datasets, and ImageNet-100 \cite{Tian2020multiview} and ImageNet-1K \cite{krizhevsky2012imagenet} for large datasets. Note, ImageNet-100 contains 100 classes that are randomly selected from  ImageNet-1K \cite{krizhevsky2012imagenet} and we choose the same classes with \cite{Tian2020multiview}. For STL-10, both 5k labeled and 100k unlabeled images are used for the pretrained model, and only 5k labeled images are used for the linear evaluation.

\textbf{Augmentation:} In this paper, we define "aggressive" augmentation including random crop, horizontal flip, grayscale, color jitter and Gaussian blur, while "weak" augmentation includes random crop and horizontal flip. The hyper-parameters of the augmentations are following MoCo v2 \cite{he2020mocov2} except the size of the cropped images for the small and medium datasets, and we resize images to $32 \times 32$ and $64 \times 64$ for the small and medium datasets, respectively.

\textbf{Baselines:} For the comparison, we re-implement the state-of-the-art methods, SimCLR \cite{Chen2020simclr}, MoCo v2 \cite{he2020mocov2}, SimSiam \cite{he2021simsiam} and BYOL \cite{Grill2020byol} based on the public codes. We follow \cite{he2021simsiam} that implements the MoCo v2 with a symmetrized loss function, and set the exponential moving average factor to 0.99 for all experiments. For BYOL \cite{Grill2020byol} on the small and medium datasets, we follow \cite{Grill2020byol}, and set the channel inner layer of 1024 in the projection and prediction MLP and the output feature is 128. We set the exponential moving average factor beginning from 0.99 with a slowing increase to 1.  

\textbf{Network structure and optimization:} Our method and reproduced methods are implemented by PyTorch v1.8 and we conduct all experiments on Nvidia V100. Our method is based on our reproduced BYOL \cite{Grill2020byol}. The code of the proposed method is in the \textit{supplementary material} and will be published after the paper is accepted. 

For the pre-train stage on the small and medium datasets, we adopt ResNet-18 \cite{He2016ResNet} as backbone. For optimization, we use SGD optimizer with a a cosine-annealed learning rate of 0.1 \cite{Loshchilov2017CosineLr}, and momentum 0.9, and a weight decay of $5 \times 10^{-4}$ and a batch size of 256. We set $\beta_{base}=0.3$ for CIFAR-10/100, and $\beta_{base}=0.4$ for Tiny ImageNet and STL-10. 

For the pre-train stage on large datasets, we adopt a standard ResNet-50 \cite{He2016ResNet} as backbone. For ImageNet-100, the network is trained using  SGD optimizer with a single cycle of cosine annealing \cite{Loshchilov2017CosineLr}, and initial learning rate of 0.2, and momentum 0.9, and a weight decay of $10^{-4}$, and a batch size of 256. We conduct ImageNet-1K experiments with $8 \times $ Nvidia V100 32G with Automatic Mixed Precision (AMP) package \cite{Paulius18amp}. Specifically, we follow \cite{Grill2020byol}, and train a network with a LARS optimizer \cite{Yang2017LARS} with a single cycle of cosine annealing \cite{Loshchilov2017CosineLr}, and a momentum of 0.9, and a weight decay of $10^{-6}$, and a batch size of 2048. The base learning rate starts from 0.9 and 0.6 for 100 and 200 epochs respectively, linearly scaled by the times of batch size 256 \cite{Goyal20171Hour}. We set $\beta_{base}=0.4$ for ImageNet-100 and ImageNet-1K.

\textbf{Evaluation:} We evaluate the representations of the pre-trained model with the linear evaluation protocol, which freezes the encoder parameters and trains a linear classifier on top of the pre-trained model. For the small and medium datasets, we follow the setting in MoCo v2 \cite{he2020mocov2} and train a linear classifier for 100 epochs with an initial learning rate of 30, no weight decay, and a momentum of 0.9. The learning rate will be multiplied by 0.1 at the 60 and 80 epochs. For the large datasets, we follow the evaluation setting in Mean Shift \cite{Soroush20P21meanshift}, which only requires 40 epochs and a batch size of 256. The linear classifier is trained with SGD and an initial learning rate of 0.01, weight decay of $10^{-4}$ and a momentum of 0.9. The learning rate will be multiplied by 0.1 at 15, 30 and 40 epochs.

\subsection{Preliminary Analysis} 

\begin{table}[!tp]
\fontsize{9}{10}\selectfont
\centering
\caption{Analysis of noise impacts of aggressive augmentations. We run methods on small and medium datasets for 200 epochs, and on ImageNet-100 for 100 epochs. The mean and standard deviation are computed over three trails.}
\vspace{5pt}
\scalebox{1}{
\begin{tabular}{c | c | c | c | c | c | c | c } 
\Xhline{2\arrayrulewidth}
	Method          & Aug.   & $\beta$  & CIFAR-10       & CIFAR-100    & STL-10          & Tiny ImageNet    & ImageNet-100   \\ \hline
	BYOL            & AA     & --     & 90.3$\pm$0.1   & 64.7$\pm$0.3 & 91.6$\pm$0.1    & 49.7$\pm$0.4     & 80.9           \\
	BYOL            & AW     & --     & 90.7$\pm$0.1   & 66.0$\pm$0.4 & 90.7$\pm$0.3    & 51.4$\pm$0.2     & 81.2           \\
    MSR             & MA   & Fixed  & 91.5$\pm$0.2   & 67.2$\pm$0.1 & \textbf{93.0$\pm$0.1}    & 53.6$\pm$0.3     & 83.2           \\
    MSR             & MA   & Decay  & \textbf{92.2$\pm$0.1}   & \textbf{68.1$\pm$0.0} & \textbf{93.0$\pm$0.1}    & \textbf{54.4$\pm$0.1}     & \textbf{83.5}           \\

\Xhline{2\arrayrulewidth}
\end{tabular}
}
\label{tab:prelim}
\vspace{-5pt}
\end{table}

In this subsection, we investigate the influence of noise from aggressive augmentations on five datasets. We compare the linear performance of BYOL with two aggressive augmentations (AA), and with one aggressive and one weak augmentation (AW). Table \ref{tab:prelim} presents that BYOL with AW can largely improve the performance on CIFAR-100 and Tiny ImageNet, and mildly improve the performance on CIFAR-10 and ImageNet-100, but in STL, the network suffers from the weak augmentation. This inconsistent results verify our claim that aggressive augmentations have different degree of detrimental effects on different datasets. For those with more number of classes and small size images, it will have more negative impacts. By contrast, using weak augmentation may result in suboptimal results for datasets with large images and fewer classes.

Then, we compare BYOL with our proposed method MSR. In the third row in Table \ref{tab:prelim}, we report the results of MSR with multi-stage augmentation (MA) and fixed $\beta$. MSR has improved linear classification across across five datasets, suggesting that aggressive augmentations can be leveraged while balancing negative effects. The results in the fourth row continue to improve by reducing $\beta$ during the training process, which indicates that noise impacts vary at different stages of the training process and will become more apparent at the end. Notably, we set $\beta_{base}$ as a default number of 0.5 in preliminary experiments, which means the weights of aggressive augmented instances and weak augmented are equal. Compared with Table \ref{tab:small} and \ref{tab:imagenet100}, we can see the performance can be further improved with a turned $\beta_{base}$.

\subsection{Linear Classification}

\begin{table}[!tp]
\fontsize{9}{10}\selectfont
\centering
\caption{Performance comparison with linear classification on small and medium datasets for 200 and 800 epochs. We adopt a ResNet-18 as backbone for all experiments. The mean and standard deviation are computed over three trails.}
\vspace{5pt}
\scalebox{1}{
\begin{tabular}{c | c | c | c | c | c | c | c | c } 
\Xhline{2\arrayrulewidth}
	\multirow{2}{*}{Method}  & \multicolumn{2}{c |}{CIFAR-10}   & \multicolumn{2}{c |}{CIFAR-100} & \multicolumn{2}{c |}{STL-10} & \multicolumn{2}{c }{Tiny ImageNet} \\ \cline{2-9}
	                 & 200 ep       & 800 ep   & 200 ep       & 800 ep  & 200 ep       & 800 ep & 200 ep       & 800 ep      \\ \hline
	SimCLR           & 88.5$\pm$0.1 & 91.3     & 60.7$\pm$0.6 & 64.4    & 87.9$\pm$0.4 & 91.1   & 46.6$\pm$0.1 & 49.1        \\
	MoCo v2          & 86.9$\pm$0.2 & 90.8     & 60.5$\pm$0.3 & 65.0    & 88.3$\pm$0.4 & 91.2   & 47.6$\pm$0.2 & 50.7        \\
	SimSiam          & 87.6$\pm$0.2 & 91.6     & 56.2$\pm$1.2 & 62.3    & 85.7$\pm$0.3 & 89.7   & 41.0$\pm$0.3 & 43.9        \\
	BYOL             & 90.3$\pm$0.1 & 92.5     & 64.7$\pm$0.3 & 69.5    & 91.6$\pm$0.1 & 93.6   & 49.7$\pm$0.4 & 53.4        \\ \hline
	MSR (Ours)            & \textbf{92.4$\pm$0.1} & \textbf{93.9}  & \textbf{68.9$\pm$0.3} & \textbf{71.3}   & \textbf{93.3$\pm$0.2} & \textbf{94.3} & \textbf{54.9$\pm$0.2} & \textbf{56.7} \\
\Xhline{2\arrayrulewidth}
\end{tabular}
}
\label{tab:small}
\vspace{-5px}
\end{table}

\textbf{Small and medium datasets.} We first verify the effectiveness of our method on small and medium datasets, with short training time, 200 epochs and long training time, 800 epochs. We also run three trails for short running time to evaluate stability of the proposed method. Table \ref{tab:small} illustrates that the proposed method significantly improve the performance across the four datasets on short training time experiments, demonstrating that MSR can accelerate convergence and the small standard deviation also shows that the proposed method has good stability. For the long training time experiments, outstanding results show MSR can continue to improve the final performance.

\begin{wraptable}{r}{20em}
\fontsize{9}{10}\selectfont
\centering
\vspace{0pt}
\caption{Performance comparison with linear classification on ImageNet-100. All methods adopt ResNet-50 as backbone. }
\vspace{5pt}
\scalebox{1}{
\begin{tabular}{c | c | c | c } 
\Xhline{2\arrayrulewidth}
	Method           & Batch size & 100 ep         & 200 ep      \\ \hline
	SimCLR           & 256        & 79.1           & 82.4            \\
	MoCo v2          & 256        & 80.9           & 83.9            \\
	SimSiam          & 256        & 79.7           & 82.6            \\
	BYOL             & 256        & 80.9           & 83.6            \\ \hline
	MSR (Ours)       & 256        & \textbf{83.7}  & \textbf{85.5}   \\
\Xhline{2\arrayrulewidth}
\end{tabular}
} 
\label{tab:imagenet100}
\vspace{-5px}
\end{wraptable}

\begin{table}[!tp]
\fontsize{9}{10}\selectfont
\centering
\vspace{0pt}
\caption{Performance comparison with linear classification on ImageNet-1K. All methods use a standard ResNet-50 as backbone without multi-crop strategy.}
\vspace{5pt}
\scalebox{1}{
\begin{tabular}{c | c | c | c | c  } 
\Xhline{2\arrayrulewidth}
	Method                          & Neg. pairs   & Batch Size & Epochs        & Top-1 Linear  \\ \hline
	Supervised                      &              & 256        & 100           & 76.2          \\ 
	InstDis \cite{Wu2018instance}   & \checkmark   & 256        & 200           & 56.5          \\
    PIRL    \cite{Misra2020Invariant}& \checkmark  & 256        & 200           & 63.6          \\
    SimCLR  \cite{Chen2020simclr}   & \checkmark   & 4096       & 1000          & 69.3          \\
	MoCo v2 \cite{he2020mocov2}     & \checkmark   & 256        & 200           & 67.5          \\
	JCL     \cite{Qi2020JCL}        & \checkmark   & 256        & 200           & 68.7          \\ 
	ReSSL   \cite{Mingkai2021ressl} & \checkmark   & 256        & 200           & 69.6          \\ 
	InfoMin Aug. \cite{Tian2020GoodViews}& \checkmark   & 256        & 200           & 70.1          \\ 
	W-MSE 4 \cite{Ermolov2021whiten}&              & 256        & 400           & 72.6          \\  
    SimSiam \cite{he2021simsiam}    &              & 256        & 200           & 70.0          \\
	SwAV    \cite{he2021simsiam}    &              & 4096       & 200           & 69.1          \\
	BYOL    \cite{he2021simsiam}    &              & 4096       & 100           & 66.5          \\
	BYOL    \cite{he2021simsiam}    &              & 4096       & 200           & 70.6          \\
	BYOL    \cite{Grill2020byol}    &              & 4096       & 300           & 72.6          \\\hline
	MSR (Ours)                      &              & 2048       & 100           & 71.4          \\
	MSR (Ours)                      &              & 2048       & 200           & \textbf{73.1} \\
\Xhline{2\arrayrulewidth}
\end{tabular}
}
\label{tab:imagenet}
\vspace{-5px}
\end{table}

\textbf{Large datasets.} We evaluate the performance of the proposed method on the large datasets, ImageNet-100 and ImageNet-1K. We reproduce all baselines on ImageNet-100 with batch size 256. The results on ImageNet-100 and ImageNet-1K are shown in Table \ref{tab:imagenet100} and Table \ref{tab:imagenet} respectively. We can see that MSR outperforms the state-of-the-art methods on ImageNet-100 with a relatively large margin across 100 and 200 epochs. For results on ImageNet-1K, MSR consistently surpasses baselines, e.g., the performance of MSR for 100 epochs has already surpassed BYOL training for 200 epochs. MSR achieves a new state-of-the-art result for 200 epochs, exhibiting a 2.5\% improvement over BYOL. Overall, empirical results on linear evaluation verify that MSR can accelerate the convergence rate and improve the generalization on various settings.

\begin{table}[!tp]
\fontsize{9}{10}\selectfont
\centering
\caption{\textbf{Transfer learning on downstream tasks:} object detection, instance segmentation and keypoint detection. All models pretrained on ImageNet-1K for 200 epochs and fine-tuned on MS COCO with 1 × schedule. Object detection and instance segmentation results are from \cite{Tian2020GoodViews} and \cite{Wang2021uota} and  keypoint detection results are from \cite{Wang2021uota}. Results with * uses multi-crop strategy.}
\vspace{5pt}
\scalebox{1}{
\begin{tabular}{c | c | c | c | c | c | c | c | c | c } 
\Xhline{2\arrayrulewidth}
	\multirow{2}{*}{Method}   & \multicolumn{3}{c |}{Object detection}          & \multicolumn{3}{c |}{Instance segmentation}     & \multicolumn{3}{c }{Keypoint detection}        \\ \cline{2-10}
	            & $AP^{bb}$  & $AP^{bb}_{50}$   & $AP^{bb}_{75}$  & $AP^{mk}$  & $AP^{mk}_{50}$   & $AP^{mk}_{75}$  & $AP^{kp}$  & $AP^{kp}_{50}$   & $AP^{kp}_{75}$ \\ \hline
	Random      & 32.8       & 50.9             & 35.3            & 29.9       & 47.9             & 32.0            & 63.5       & 85.3             & 69.3            \\
	Supervised  & 39.7       & 59.5             & 43.3            & 35.9       & 56.6             & 38.6            & 65.4       & 87.0             & 71.0            \\
	MoCo \cite{he2020moco}         & 39.4       & 59.1             & 42.9            & 35.1       & 55.9             & 37.7            & 65.6       & 87.1             & 71.3            \\
	MoCo v2 \cite{he2020mocov2}    & 40.1       & 59.8             & 44.1            & 35.3       & 55.9             & 37.9            & 66.0       & 87.2             & 71.4            \\
	InfoMin Aug. \cite{Tian2020GoodViews}& 40.6       & 60.6             & 44.6            & 36.7       & 57.7             & 39.4            & --.-       & --.-             & --.-            \\
	JCL \cite{Qi2020JCL}           & --.-       & --.-             & --.-            & 35.6       & 56.2             & 38.3            & 66.2       & 87.2             & \textbf{72.3}            \\
	SwAV* \cite{Caron2020SwAV}     & --.-       & --.-             & --.-            & 36.3       & 57.7             & 38.9            & 65.6       & 86.9             & 71.6            \\     
	UOTA* \cite{Wang2021uota} & --.-       & --.-             & --.-            & 36.7       & 58.4             & 39.4            & \textbf{66.3}       & \textbf{87.4}    & \textbf{72.3}            \\ \hline
	MSR (Ours)  & \textbf{41.1} & \textbf{61.4} & \textbf{45.1}   & \textbf{37.3} & \textbf{58.6} & \textbf{40.1}   & 66.1       & \textbf{87.4}    & 72.0            \\
\Xhline{2\arrayrulewidth}
\end{tabular}
}
\label{tab:coco}
\vspace{-5px}
\end{table}

\begin{table}[!tp]
\fontsize{9}{10}\selectfont
\centering
\caption{\textbf{Transfer learning on downstream tasks:} object detection, and instance segmentation. All models pretrained on ImageNet-1K for 200 epochs and fine-tuned on MS COCO with 2 × schedule. Baseline results are from \cite{Tian2020GoodViews}.}
\vspace{5pt}
\setlength{\tabcolsep}{4.5mm}{
\begin{tabular}{c | c | c | c | c | c | c } 
\Xhline{2\arrayrulewidth}
	\multirow{2}{*}{Method}   & \multicolumn{3}{c |}{Object detection}          & \multicolumn{3}{c }{Instance segmentation}  \\ \cline{2-7}
	            & $AP^{bb}$  & $AP^{bb}_{50}$   & $AP^{bb}_{75}$  & $AP^{mk}$  & $AP^{mk}_{50}$   & $AP^{mk}_{75}$  \\ \hline
	Random      & 38.4       & 57.5             & 42.0            & 34.7       & 54.8             & 37.2            \\
	Supervised  & 41.6       & 61.7             & 45.3            & 37.6       & 58.7             & 40.4            \\
	InstDis \cite{Wu2018instance}    & 41.3       & 61.0             & 45.3            & 37.3       & 58.3             & 39.9            \\
	PIRL \cite{Misra2020Invariant}       & 41.2       & 61.2             & 45.2            & 37.4       & 58.5             & 40.3            \\
	MoCo \cite{he2020moco}      & 41.7       & 61.4             & 45.7            & 37.5       & 58.6             & 40.5            \\
	MoCo v2 \cite{he2020mocov2} & 41.7       & 61.6             & 45.6            & 37.6       & 58.7             & 40.5            \\
	InfoMin Aug. \cite{Tian2020GoodViews}& 42.5       & 62.7             & 46.8            & 38.4       & 59.7             & \textbf{41.4}            \\ \hline
	MSR (Ours)  & \textbf{42.7}       & \textbf{62.9}             & \textbf{47.1}            & \textbf{38.5}   & \textbf{60.0} & \textbf{41.4}            \\ 
\Xhline{2\arrayrulewidth}
\end{tabular}
}
\label{tab:coco2}
\vspace{-5px}
\end{table}

\subsection{Transfer Learning}
We further verify the quality of representation learned by MSR on more downstream tasks. For object detection and instance segmentation, we follow \cite{Tian2020GoodViews}, and adopt Mask R-CNN \cite{Kaiming2017maskrcnn} with FPN \cite{Kaiming2017fpn} to fine-tune our pretrained ResNet-50 model on COCO \textit{train2017} with 1 × schedule and 2 × schedule, and evaluate performance on COCO \textit{val2017}. Similar to object detection, we change Mask R-CNN to a keypoint version to conduct keypoint detection experiments with 1 × schedule. For more experimental details, please check the \textit{supplementary material}.

Table \ref{tab:coco} and Table \ref{tab:coco2} report the results of object detection, instance segmentation and keypoint detection. We can observe that MSR outperforms all baselines on object detection and instance segmentation tasks, especially, showing superior over the strong baseline InfoMin Aug. \cite{Tian2020GoodViews}. For the keypoint detection task, MSR also has comparable performance with UOTA, less than $0.3\%$. Note that UOTA employs multi-crop strategy with 8 views (2 × 224 + 6 × 96), which means although UOTA and MSR both train for 200 epochs, UOTA receives much more examples than MSR. Strong performance on downstream tasks demonstrates that MSR can improve the general quality of learned representation.

\subsection{Training time comparison}
\label{sec:time}

We compare the running time between our method and reproduced BYOL. We conduct experiments on CIFAR-100 and STL-10 for 800 epochs with a single Nvidia V100, ImageNet-100 for 200 epochs with $4 \times $ Nvidia V100, respectively. For ImageNet-100, we use Pytorch Automatic Mixed Precision package to speed up the training process and save GPU memory. Note that because we do not have the enough GPUs to run a standard BYOL with a batch size of 4096 on ImageNet, we use ImageNet-100 instead of ImageNet-1K, which has similar processing efficiency but ten times data.

\begin{wraptable}{r}{0.5\textwidth}
\fontsize{9}{10}\selectfont
\centering
\vspace{-5px}
\caption{Training time comparison with BYOL on three datasets}
\vspace{5px}
\begin{tabular}{c | c | c | c } 
\Xhline{2\arrayrulewidth}
	Method  & CIFAR-100 & STL-10 & ImageNet-100 \\ \hline
	BYOL    & 11.3h     & 77.6h  & 9.5h         \\
	MSR     & 11.5h     & 79.6h  & 10.4h        \\
\Xhline{2\arrayrulewidth}
\end{tabular}
\label{tab:time}
\vspace{0px}
\end{wraptable}

Although our method uses double instances in the loss function \ref{eq:loss}, each instance in MSR only passes into the network one time. Therefore, the number of forward and backward keeps the same with BYOL. In addition, thanks to the efficient multi-stage augmentation, where we generate double instances using the nearly same resources. The main burden training time part of our proposed method may come from two more vector multiplication in our loss function \ref{eq:loss}. From Table \ref{tab:time}, we can see that our method is as efficient as BYOL, and the differences between the two methods are less than 9\% across three datasets. 
\subsection{Ablation Studies}

\begin{wrapfigure}{r}{0.5\textwidth}
\centering
\vspace{0px}
\includegraphics[width=2.5in]{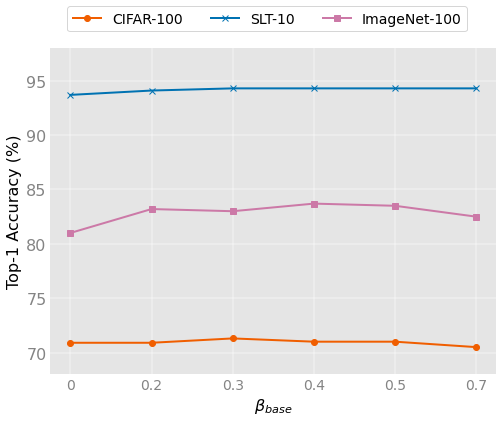}
\captionof{figure}{Sensitivity analysis for the hyper-parameter $\beta_{base}$. We conduct experiments on CIFAR-100 and STL-10 for 800 epochs, and ImageNet-100 for 100 epochs, respectively.}
\label{fig:abla}
\vspace{-5px}
\end{wrapfigure} 

In this section, we investigate the sensitivity of the hyper-parameter $\beta_{base}$ on three datasets. The linear classification results are shown in Figure \ref{fig:abla}. First of all, we can observe that MSR is not very sensitive to the value of $\beta_{base}$. MSR with $\beta_{base}=0.3$ achieves the highest accuracy on CIFAR-100, and the best results on STL-10 and ImageNet-100 occur at $\beta_{base}=0.4$.  The performance with aggressive and weak augmented instances on the target network is higher than that with only weak augmented instances ($\beta_{base}=0$), which suggests that only employing weak augmented samples on the target network may lead to suboptimal performance.

Furthermore, the value of $\beta_{base}$ on medium and large datasets is higher than that on small datasets, which indicates that small datasets require higher weights to weak augmented samples to against noise impacts from aggressive augmented samples. This is consistent with our empirical finds that small datasets have higher probabilities of generating noise than medium and large datasets when employing the same strategy of aggressive augmentations.

\section{Conclusion}
In this paper, we first empirically demonstrate that two positive instances generated by the aggressive augmentations can cause the semantic shift issue, which introduces noisy positive pairs and degrades the quality of learned representation. To alleviate this issue, we propose a novel method Making Self-supervised learning Robust (MSR) to properly utilize aggressive augmentations and neutralize the semantic shift problem. Experimental results show that our proposed method achieves state-of-the-art results with the linear evaluation on various datasets and consistently improves the generalization on a series of downstream tasks.

The main limitation of this paper is that although we know there is noise from aggressive augmentations, we are still unknown the specific conditions under which noise occurs and how much networks suffer from it. We leave it as the future research direction.

\section*{Acknowledgments}\label{sec:6}
The authors would give special thanks to ...

\bibliographystyle{plainnat}
\bibliography{ref}

\newpage
\appendix
\onecolumn

\section{Implementation Details}

\subsection{Object detection and instance segmentation}

Following the common settings used in \cite{he2020moco, Tian2020GoodViews}, we employ the Mask R-CNN \cite{Kaiming2017maskrcnn} with FPN \cite{Kaiming2017fpn} based on \textit{Detectron2} \cite{wu2019detectron2}.
We fine-tune the pre-trained model on COCO \textit{train2017} and evaluate it on COCO \textit{val2017} \cite{Lin2014COCO} with $8 \times $ Nvidia V100.
For the 1 $\times$ schedule experiment, the learning rate is initialized as $0.03$. For the 2 $\times$ schedule experiment, we set the initial learning rate as $0.02$.

\subsection{Keypoint detection}

Following \cite{Wang2021uota}, we adopt a keypoint version Mask R-CNN \cite{Kaiming2017maskrcnn} with FPN \cite{Kaiming2017fpn} based on \textit{Detectron2} \cite{wu2019detectron2}.
We fine-tune the pre-trained model on COCO \textit{train2017} with 1 $\times$ schedule and evaluate it on COCO \textit{val2017} \cite{Lin2014COCO}.
We conduct experiments with $8 \times $ Nvidia V100, and the initial learning rate is set as $0.05$.

\section{Additional experiments}

\begin{table}[h]
\centering
\caption{Performance comparison with linear classification without the fixing learning rate trick on three datasets for 200 epochs. The mean and standard deviation are computed over three trails.}
\vspace{5pt}
\begin{tabular}{c | c | c | c } 
\Xhline{2\arrayrulewidth}
	Method           & CIFAR-100                & STL-10         & ImageNet-100            \\ \hline
	SimSiam          & 56.2$\pm$1.2             & 85.7$\pm$0.3   & 82.6                    \\ 
	Simsiam + MSR    & \textbf{63.8$\pm$0.4}    & \textbf{89.9$\pm$0.5}  & \textbf{84.0}           \\

\Xhline{2\arrayrulewidth}
\end{tabular}
\label{tab:fixinglr}
\end{table}

To evaluate the adaptability, we build our proposed MSR on an additional self-supervised framework, SimSiam \cite{he2021simsiam}, termed SimSiam + MSR. Specifically, we change the same similarity function from mean square error to negative cosine similarity. we adopt the same settings for all the hype-parameters mentioned in Section 4.1  and conduct a series of experiments on CIFAR-100 \cite{krizhevsky2009CIFAR}, STL-10 \cite{Adam2011stl10}, and ImageNet-100 \cite{Tian2020multiview} for 200 epochs. 

As illustrated in Table~\ref{tab:fixinglr}, we observe that the proposed MSR achieves better transfer performance than Simsiam on three datasets. For instance, MSR significantly raise the accuracy of linear probing from 56.2\% to 63.8\% (+7.6\%) on the CIFAR-100 dataset. Therefore, our proposed method does not rely on the momentum network, demonstrating the adaptability of MSR.

\end{document}